\pdfoutput=1
\documentclass{article}
\usepackage[utf8]{inputenc} %
\usepackage[T1]{fontenc}    %

\usepackage[bitstream-charter]{mathdesign}
\usepackage{amsmath}
\usepackage[scaled=0.92]{PTSans}

\usepackage[
  paper  = letterpaper,
  top    = 1.0in,
  bottom = 1.0in,
  ]{geometry}

\usepackage[usenames,dvipsnames,table]{xcolor}
\definecolor{shadecolor}{gray}{0.9}

\usepackage[final,expansion=alltext]{microtype}
\usepackage[english]{babel}
\usepackage[parfill]{parskip}
\usepackage{afterpage}
\usepackage{framed}

{\endMakeFramed}

\usepackage{lineno}

\usepackage{ragged2e}

\newcommand{\parnum}{\bfseries\P\arabic{parcount}}
\newcounter{parcount}
\newcommand\p{%
    \stepcounter{parcount}%
    \leavevmode\marginpar[\hfill\parnum]{\parnum}%
}
\usepackage{graphicx}
\usepackage{wrapfig}
\usepackage[labelfont=bf,font=footnotesize,width=.9\textwidth]{caption}
\usepackage[format=hang]{subcaption}

\usepackage{booktabs,multirow,multicol}       %

\usepackage[algoruled,algo2e]{algorithm2e}
\usepackage{listings}
\usepackage{fancyvrb}
\fvset{fontsize=\normalsize}

\usepackage[colorlinks,linktoc=all]{hyperref}
\usepackage[all]{hypcap}
\hypersetup{citecolor=MidnightBlue}
\hypersetup{linkcolor=MidnightBlue}
\hypersetup{urlcolor=MidnightBlue}

\usepackage[nameinlink, capitalize]{cleveref}

\usepackage[acronym,smallcaps,nowarn]{glossaries}[=v4.46]
\lstdefinestyle{mystyle}{
    commentstyle=\color{OliveGreen},
    keywordstyle=\color{BurntOrange},
    numberstyle=\tiny\color{black!60},
    stringstyle=\color{MidnightBlue},
    basicstyle=\ttfamily,
    breakatwhitespace=false,
    breaklines=true,
    captionpos=b,
    keepspaces=true,
    numbers=left,
    numbersep=5pt,
    showspaces=false,
    showstringspaces=false,
    showtabs=false,
    tabsize=2
}
\usepackage{amsmath,amsfonts,bm}

\def\eqref#1{equation~\ref{#1}}

\def\1{\bm{1}}

\def\mP{{\bm{P}}}
\def\mQ{{\bm{Q}}}

\def\mW{{\bm{W}}}
\def\mX{{\bm{X}}}

\DeclareMathAlphabet{\mathsfit}{\encodingdefault}{\sfdefault}{m}{sl}
\SetMathAlphabet{\mathsfit}{bold}{\encodingdefault}{\sfdefault}{bx}{n}

\def\gG{{\mathcal{G}}}

\newcommand{\E}{\mathbb{E}}

\newcommand{\R}{\mathbb{R}}

\usepackage{csquotes}
\usepackage[%
minnames=1,maxnames=99,maxcitenames=2,
style=alphabetic,
doi=false,
url=false,
giveninits=true,
hyperref,
natbib,
backend=bibtex,
sorting=nyt,
backref=true
]{biblatex}%
\renewbibmacro{in:}{%
  \ifentrytype{article}{}{\printtext{\bibstring{in}\intitlepunct}}}

\renewbibmacro*{journal}{%
  \iffieldundef{journaltitle}
    {}
    {\printtext[journaltitle]{%
       \printfield[noformat]{journaltitle}%
       \setunit{\subtitlepunct}%
       \printfield[noformat]{journalsubtitle}}}}

\DeclareFieldFormat{sentencecase}{\MakeSentenceCase*{#1}}

\renewbibmacro*{title}{%
  \ifthenelse{\iffieldundef{title}\AND\iffieldundef{subtitle}}
    {}
    {\ifthenelse{\ifentrytype{article}\OR\ifentrytype{inbook}%
      \OR\ifentrytype{incollection}\OR\ifentrytype{inproceedings}%
      \OR\ifentrytype{inreference}\OR\ifentrytype{misc}}
      {\printtext[title]{%
        \printfield[sentencecase]{title}%
        \setunit{\subtitlepunct}%
        \printfield[sentencecase]{subtitle}}}%
      {\printtext[title]{%
        \printfield[titlecase]{title}%
        \setunit{\subtitlepunct}%
        \printfield[titlecase]{subtitle}}}%
     \newunit}%
  \printfield{titleaddon}}

\AtEveryBibitem{%
\ifentrytype{article}{
    \clearfield{urldate}%
    \clearfield{eprint}
    \clearfield{eid}
}{}
\ifentrytype{book}{
    \clearfield{url}%
    \clearfield{urldate}%
    \clearfield{eprint}
}{}
\ifentrytype{collection}{
    \clearfield{url}%
    \clearfield{urldate}%
    \clearfield{eprint}
}{}
\ifentrytype{incollection}{
    \clearfield{url}%
    \clearfield{urldate}%
    \clearfield{eprint}
}{}
}

\AtEveryBibitem{
    \clearfield{pages}
    \clearfield{review}%
    \clearfield{series}%
    \clearfield{volume}
    \clearfield{month}
    \clearfield{isbn}
    \clearfield{issn}
    \clearlist{location}
    \clearfield{series}
    \clearlist{publisher}
    \clearname{editor}
}{}

\usepackage{times}
\usepackage{hyperref}
\usepackage{url}
\usepackage{graphicx}
\usepackage{subcaption}
\usepackage{placeins}
\usepackage{subcaption}
\usepackage{algorithm,algpseudocode}
\usepackage{xcolor}
\usepackage{float}
\usepackage{enumitem}

\newcommand{\new}

\usepackage{xcolor}

\definecolor{WowColor}{rgb}{.75,0,.75}
\definecolor{SubtleColor}{rgb}{0,0,.50}

\newcounter{margincounter}

\usepackage[affil-it]{authblk}

\usepackage{makecell}

\usepackage{multirow}
\usepackage{tikz}
\usetikzlibrary{positioning}
\usetikzlibrary{arrows, automata}
\usetikzlibrary{patterns}
\def\showauthornotes{1}
\ifnum\showauthornotes=1
\newcommand{\Authornote}[2]{{\sf\small\color{blue}{[#1: #2]}}}
\else
\newcommand{\Authornote}[2]{}
\fi

\title{Exploring the Manifold of Neural Networks\\Using Diffusion Geometry}

\author{
\begin{center}
Elliott Abel\textsuperscript{1}\thanks{co-first authors}~~Andrew J. Steindl\textsuperscript{5}$^{\ast}$~~Selma Mazioud\textsuperscript{2}$^{\ast}$~~Ellie Schueler\textsuperscript{4}$^{\ast}$~~Folu Ogundipe\textsuperscript{2}$^{\ast}$
\linebreak
Ellen Zhang\textsuperscript{2}~~~Yvan Grinspan\textsuperscript{3}~~~Kristof Reimann\textsuperscript{2}~~~Peyton Crevasse\textsuperscript{2}
\linebreak
Dhananjay Bhaskar\textsuperscript{2}~~~Siddharth Viswanath\textsuperscript{2}~~~Yanlei Zhang\textsuperscript{6}
\linebreak
Tim G. J. Rudner\textsuperscript{7}~~~Ian Adelstein\textsuperscript{2}~~~Smita Krishnaswamy\textsuperscript{2}
\linebreak\linebreak
$^{1}$Pomona College~~~~$^{2}$Yale University~~~~$^{3}$Amherst College~~~~$^{4}$Bates College
\linebreak
$^{5}$Vassar College~~~~$^{6}$Mila - Quebec AI Institute~~~~$^{7}$New York University
\end{center}
}

\date{}

\begin{document}

\maketitle

\begin{abstract}
Drawing motivation from the manifold hypothesis, which posits that most high-dimensional data lies on or near low-dimensional manifolds, we apply manifold learning to the space of neural networks. We learn manifolds where datapoints are {\em neural networks} by introducing a distance between the hidden layer representations of the neural networks. These distances are then fed to the non-linear dimensionality reduction algorithm PHATE to create a manifold of neural networks. We characterize this manifold using features of the representation, including class separation, hierarchical cluster structure, spectral entropy, and topological structure. Our analysis reveals that high-performing networks cluster together in the manifold, displaying consistent embedding patterns across all these features. Finally, we demonstrate the utility of this approach for guiding hyperparameter optimization and neural architecture search by sampling from the manifold.
\end{abstract}

\section{Introduction}
\label{sec:intro}

The development of neural networks has revolutionized various fields by allowing models to learn intricate relationships, patterns, and underlying structures directly from data \citep{LeCun_2015, He2015DeepRL, Shelhamer2017, Szegedy2016}. However, understanding how different network architectures represent and process information, particularly in high-dimensional spaces, remains under-explored. This challenge grows more relevant as we design larger neural networks for which finding architectures and hyperparameters becomes increasingly taxing. Here we hypothesize that neural networks can be largely characterized by the data representations they create. We therefore create a manifold or space of neural networks organized by their hidden layer representations of a dataset. We show that well-performing neural networks, regardless of hyperparameter selections, lead to similar representations  \citep{Zhang2021, Papyan2020, Montavon2018}.

Our approach first collects point cloud representations of $n$ datapoints from hidden layers of a collection of neural networks and represents each point cloud as a diffusion matrix. In such a matrix, each data point is characterized by its transition probability to other data points. This gives us a representation of each neural network as an $n\times n$ diffusion operator, independent of architectural variations in hidden layer dimensions. Then to define a distance between two such neural networks, we utilize the Frobenius norm between the respective diffusion matrices. We pass these distances to PHATE \citep{moon_visualizing_2019} (Potential of Heat-diffusion for Affinity-based Transition Embedding), which embeds the neural networks into a lower-dimensional space organized by their hidden representations of the data. This mapping allows us to uncover features of the neural networks' hidden layer representations that correlate with network performance, accuracy, and other key metrics. We are then able to explore the space of neural networks in a more structured and interpretable way.

When characterizing this manifold by test performance of the neural networks, we observe contiguous regions of the manifold space that are occupied by high-performing networks. This implies that high-performing neural networks have similar hidden layer representations. We found that high-performing networks were located closer to each other in the manifold than in alternative embedding methods by computing their normalized average pairwise distance in PHATE embeddings. To understand the source of this similarity, we compare these representations by their 1) class separation, 2) hierarchical cluster structure, 3) diffusion spectral entropy, and 4) persistence homology when viewed as a graph. We find that high-performing networks exhibit similarity across all four categories.

To summarize, the key contributions of this paper are as follows:\vspace*{-2pt}
\begin{enumerate}[leftmargin=20pt]
\setlength\itemsep{0pt}
    \item
    We construct a manifold of neural networks based on their hidden representations of data.
    \item
    We characterize the regions of the manifold based on similarity in class separation, hierarchical cluster structure, entropy, and persistence homology.
    \item 
    We show how we can apply our method to searching for hyperparameters and architectures of neural networks based on sampling from the manifold.
\end{enumerate}
We envision this approach as a step towards improving our understanding of the interplay between neural network representations, neural network hyperparamters, and predictive performance.

\section{Background}
\label{sec:bg}

In this section, we provide background on data diffusion, diffusion maps and PHATE which we utilize for our neural network manifold construction.

\subsection{Computation of the Diffusion Operator} Given a point cloud $\mX$ containing $n$ points, the diffusion operator is computed by first calculating the pairwise distances between points and then constructing a similarity or adjacency matrix. We define a kernel function $\kappa:\mathbb{R}^d\times \mathbb{R}^d \to \mathbb{R}^+$ and use it to populate the adjacency matrix $\mW$, where each element is defined as $\mW_{ij} := \kappa(x_i,x_j)$ for all points $x_i, x_j \in \mX$. Here, we employ a Gaussian kernel to construct the adjacency matrix. The resulting weighted graph $\gG$, with its set of nodes representing observations and adjacency matrix values in $(0,1]$, forms the basis for defining the diffusion operator. The diffusion operator is given by $\mP:=\mQ^{-1}\mW$, where $\mQ$ is the degree matrix. This operator describes a random walk or diffusion on the graph, making it a fundamental tool for uncovering the structure of datasets in manifold learning.

\subsection{Diffusion Maps} Based on the diffusion operator, \citet{coifman_diffusion_2006} defines an embedding in $k$ dimensions via the first $k$ non-trivial right eigenvectors of the t-steps diffusion operator $\mP^t$ weighted by their eigenvalues. The embedding preserves the \textbf{\textit{diffusion distance}} $ DM_{\mP}(x_i,x_j):=\| (\boldsymbol{\delta_i}\mP^t - \boldsymbol{\delta_j}\mP^t)  (1/\pi) \|_2$, where $\boldsymbol{\delta}_i$ is a vector such that $(\boldsymbol{\delta}_i)_j = 1$ if $j=i$ and $0$ otherwise, and $\pi$ is the stationary distribution of $\mP$. Intuitively, $DM_{\mP}(x_i,x_j)$ considers all the $t$-steps paths between $x_i$ and $x_j$. A larger diffusion time can be seen as a low-frequency graph filter, keeping only information from low-frequency transitions such as the stationary distributions. The diffusion matrix converges to a stationary distribution $\pi_i := \mQ_{ii}/\sum_i\mQ_{ii}$ under mild assumptions, making diffusion with $t>1$ useful in denoising relationships between observations.

\subsection{PHATE}

PHATE is a diffusion-based method that preserves an information-theoretic \textbf{\textit{potential distance}} $PH_{\mP}:= \| -\log\boldsymbol{\delta}_i\mP^t + \log\boldsymbol{\delta}_j\mP^t \|_2$ (~\citet{moon_visualizing_2019}) and justifies this approach using the $\log$ transformation to prevent nearest neighbors from dominating the distances. Unlike diffusion maps, this method allows us to capture manifolds in very low dimensions (in particular, 2 or 3 for visualization). In our neural network manifold construction, we represent each neural network using a data diffusion operator, and then use a distance between the operators themselves to create a PHATE embedding of the space of neural networks. Note that PHATE also uses a diffusion operator internally which, in this manuscript, operates on "datapoints" that each represent a neural network. 

\section{Related Work}
\label{sec:related}
Our work builds on a rich body of research that examines and leverages the internal representations neural networks develop during training. For instance, \citet{gigante2019visualizing} employs PHATE to visualize how these representations evolve across network layers, track changes throughout training, and observe class-based variation in classification networks. Similarly, \citet{vanRossem2024representations} investigates how latent embeddings are affected by various hyperparameters, exploring the extent to which these traits are preserved across different network architectures.

The challenge of comparing latent embeddings has also seen considerable focus. For example, \citet{wang2018towards} explores subspace alignment techniques but finds they yield inconsistent structures when the same network is trained with different initializations. Canonical Correlation Analysis (CCA)-based methods have been explored by \citet{kornblith2019similarity}, who highlight desirable properties in similarity measures between learned representations, particularly those invariant to transformations. Persistent homology, a topological data analysis technique introduced by \citet{Carlsson_2020}, has also been applied in this domain. \citet{moor2020topological} designs an autoencoder that incorporates persistent homology changes between latent and ambient spaces within its loss function, while \citet{wayland2024mapping} leverages persistent homology to compare neural networks, creating a “multiverse” structured by differences in persistence landscapes—an approach similar to ours, which employs persistent homology along with hierarchical microstructure and diffusion spectral entropy.

Beyond empirical studies, theoretical investigations have examined neural network geometry. For example, extensions of the Johnson-Lindenstrauss lemma to non-linear contexts, such as fully connected and convolutional neural networks, reveal intriguing results: under certain conditions, natural image-based data maintains its geometric structure, measured by cosine similarity, while Gaussian inputs tend to exhibit contraction \citep{JLCNN}. Such findings underscore the potential for novel research into how network layers transform input data geometry, adding valuable perspective to our understanding of neural network function.

\section{Methods}
\label{sec:methods}

We introduce the following notation. Let $Y= \{ y_i\}^n_{i=1}$ be a set of $n$ test points. Throughout the manuscript, we assume that the points in $Y$ are registered, that is, that they remain consistent and in the same order throughout. Now, consider a set of $m$ neural networks, denoted $Z = \{ F_i(X, \theta_i) \}_{i=1}^{m}$, where each $F_i $ is a neural network trained on a dataset $X$, and $\theta_i$ represents the hyperparameters. For each neural network $F_i$, we define an embedding function $\phi_{F_i}$ from the input space to the hidden layer's embedding space $\mathbb{R}^{d_i}$ of network $F_i$. The hidden layer embeddings that we compare across networks are given by $\{\phi_{F_i}(Y)\}_{i=1}^m$, the collection of activations in a specific hidden layer resulting from passing the test points $Y$ into various networks $F_i$.

\subsection{The Manifold of Neural Networks}
\label{gen_inst}

Given the set of neural networks $Z$, our embedding method uses the set $Y$ of test points to obtain a representative embedding for each neural network. For a neural network $F_z \in Z$, the embedding $\phi_{F_z}(y_k)$ is extracted for each data point $y_k \in Y$. The resulting embeddings are concatenated by rows for all $k \in Y$, forming an $n \times d_z$ matrix $E_z$, where $d_z$ is the dimensionality of a hidden layer embedding for network $F_z$. The pairwise Euclidean distances between the embeddings are then computed, resulting in an $n \times n$ distance matrix $D_z$, where each entry $D_{z,ij} =  d(\phi_{F_z}(y_i), \phi_{F_z}(y_j))$.

\begin{figure*}[t!]
\centering
\includegraphics[width=.98\linewidth]{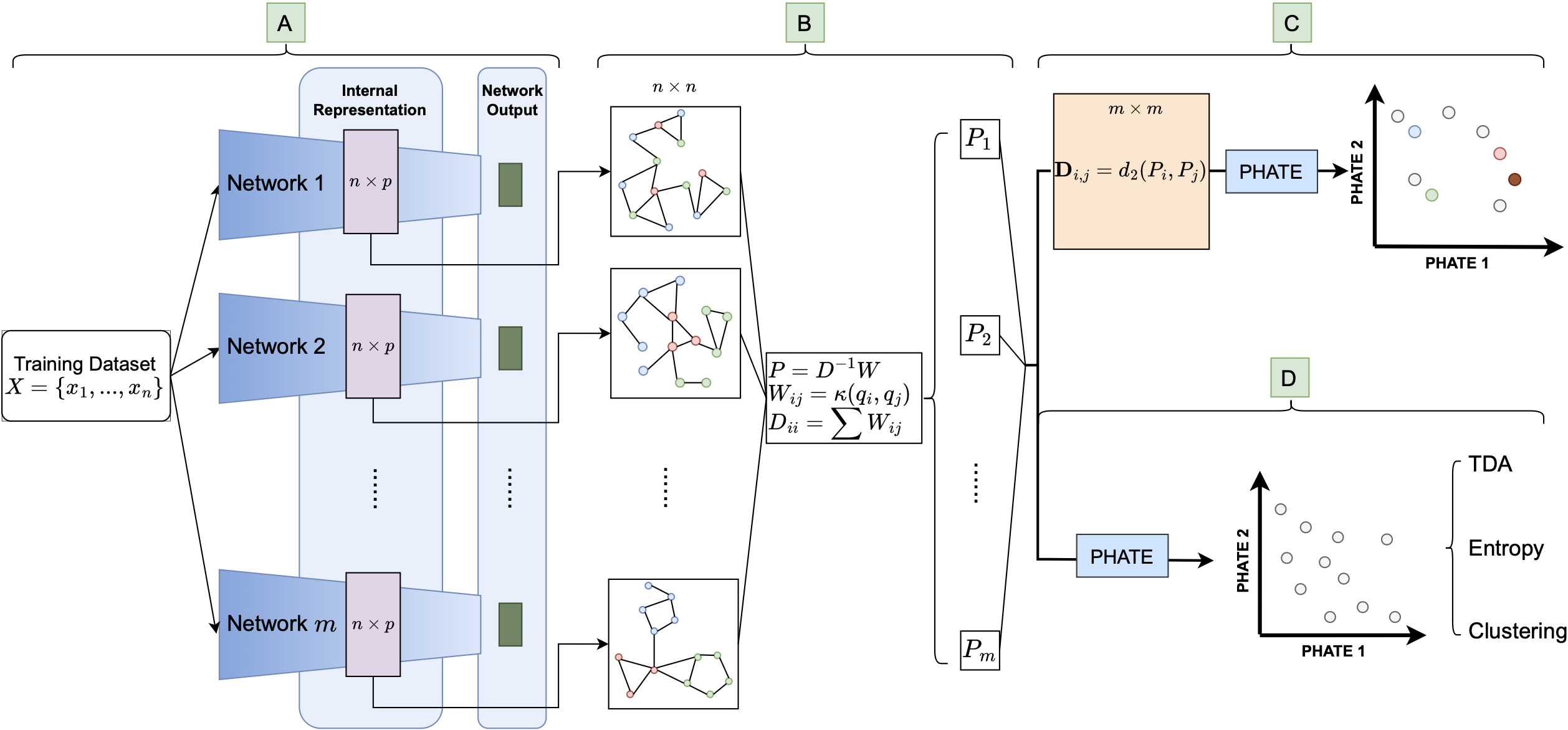}
\caption{Schematic of our method. (A) Train neural networks and retrieve their internal representation; (B) Construct graphs with edges representing pairwise diffusion probabilities; (C) Construct distance matrix of network representations and visualize network landscape for downstream tasks (D) Visualize the internal representation for characterization via topological data analysis, spectral entropy, and hierarchical clustering. }
\label{fig:schematic}
\end{figure*}

Next, $D_z$ is passed through a standard Gaussian kernel (with $\sigma = 0.5$) to obtain an $n \times n$ pairwise similarity matrix $W_z$:
\[
    W_{z,ij} = \exp \left( -\frac{d(\phi_{F_z}(y_i), \phi_{F_z}(y_j))^2}{2\sigma^2} \right),
\]
representing the edge weight between vertices $y_i$ and $y_j$ based on the distances of their hidden layer activations. The matrix $W_z$ is a fully connected graph representation of the neural network $F_z$. To ensure that the metric adheres to the underlying manifold of the data, $W_z$ is row-normalized resulting diffusion operator $P_z:= Q_z^{-1}W_z$, where $Q_{z,ij} = \delta_{ij}\sum_j W_{z,ij}$ is a continuous analog of the degree matrix.  %

This transformation from the hidden representation matrix $E_z$ to the diffusion operator $P_z$ is repeated for all neural networks $F_z \in Z$. These diffusion operators encapsulate how each neural network classifies data by describing the intrinsic geometry of their representations. Once these matrices are computed, we define the distance between neural networks as the similarity or dissimilarity of their diffusion operators $P_z$. We therefore compute the pairwise Frobenius norm between each pair of diffusion operators from the trained neural networks, resulting in the $m \times m$ distance matrix $N$ where $N_{ij} = \|P_i, P_j\|_F$. Using the non-linear dimension reduction technique PHATE to represent the distances from $N$, we are able to visualize the landscape of neural networks. This process is depicted in \Cref{fig:schematic} A-C.

\subsection{Characterizing the Landscape of Neural Networks}

\begin{figure}
\centering
\includegraphics[width=0.9\linewidth]{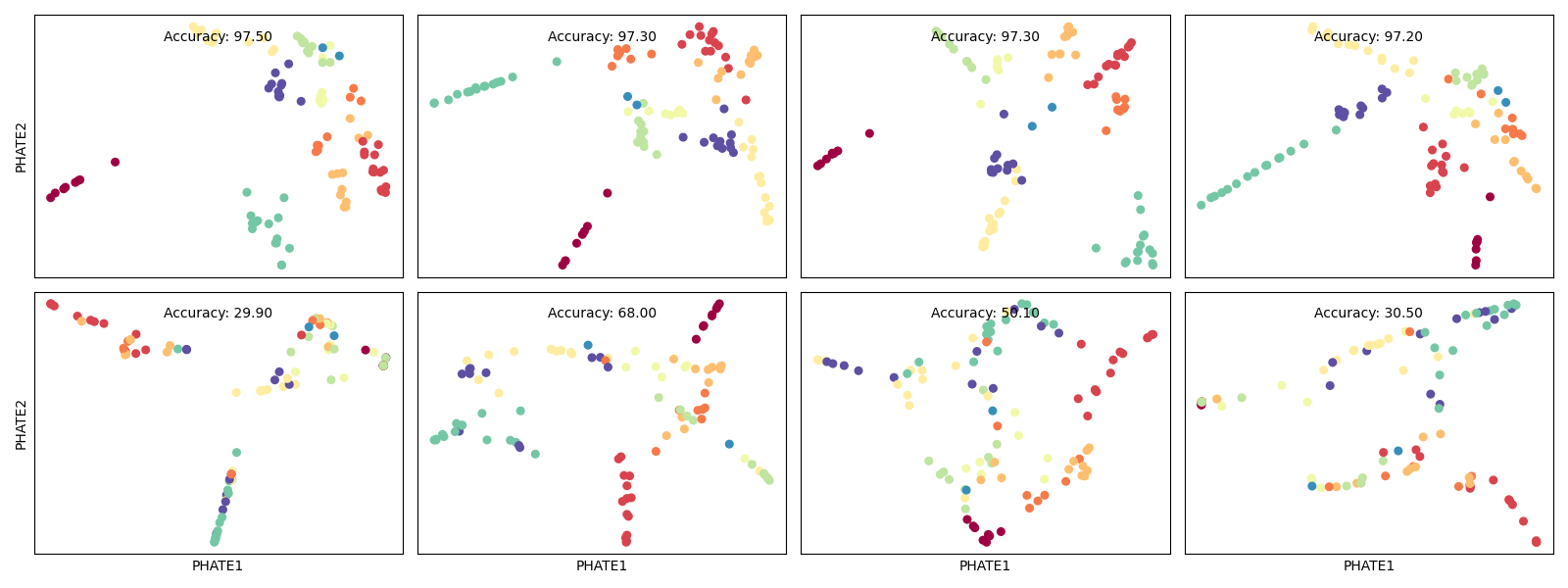}
\caption{PHATE hidden layers of neural networks trained for MNIST classification for 4 high-performing and 4 low-performing models.}
\label{fig: PHATE MLP Penultimate Layers}
\end{figure}

We characterize the manifold of neural networks over their internal representation  using class separation, hierarchical cluster microstructure, diffusion spectral entropy, and persistence homology. \Cref{fig: PHATE MLP Penultimate Layers} shows the hidden layer representations of neural networks at varying accuracies. Such attributes have strong relationships with network stability and predictive accuracy while being largely impartial to hyperparameter adjustment.

\paragraph{Class Separation and Variance.}
We characterize networks primarily by the degree of class separation within datasets that have classification labels. Intuitively, high classification accuracy requires well-separated classes, so the extent of this separation serves as a key indicator of network performance.

To quantify this separation, we measure the similarity in class distinctness by constructing a distance matrix between the centroids of each class. For a given class \( C_k \), its centroid is calculated as 
\[
\mu_k = \frac{1}{|C_k|} \sum_{x \in C_k} \phi(x),
\]
where \( \phi(x) \) represents the embedding of data point \( x \). Pairwise Euclidean distances between these centroids provide an overall sense of class structure within the representation layer. Additionally, to evaluate within-class variance, we compute the average Euclidean distance from each point in a class to its respective centroid \( \mu_k \). This within-class variance serves as a measure of compactness for each class embedding.

\paragraph{Hierarchical Clustering Structure.}
Advancing our understanding of the cluster structures formed by neural networks is essential to uncover how internal representations relate to network accuracy. We denote the hidden layer representation of \( n \) aligned data points as \( E_z \) for all \( F_z \in Z \) and disregard any true labels. Using this representation, we apply the Ward agglomerative clustering method from \citealp{Murtagh_2014}, which defines inter-cluster distances based on the increase in the sum of squares upon merging clusters.

After performing agglomerative clustering, we generate dendrograms for each model to analyze the microstructure of clusters within the network’s representations. These dendrograms offer insights into the organization of the network’s internal representations. To quantify the accuracy of these cluster structures, we compute the Adjusted Rand Index (ARI). By cutting each dendrogram at 10 clusters, we assess the similarity of each \( E_z \) clustering structure to the ground truth labels through ARI, which reveals how closely each network’s clusters align with true label groups.

Additionally, we calculate pairwise ARI values between the dendrograms of any two networks, \( F_n \) and \( F_m \), to assess similarity in their microstructures. This enables us to explore correlations between microstructure similarity and network accuracy across different networks.

\vspace*{-3pt}
\paragraph{Information Theory and Diffusion Spectral Entropy.}
The diffusion matrix of an embedding effectively captures its geometry. Looking at hidden layer embeddings of different neural network architectures, we use an information theoretic measure called Diffusion Spectral Entropy (DSE), which was introduced in \citet{liao2023assessingneuralnetworkrepresentations} to compute a fast entropy on data by first creating a data diffusion operator, and then computing the entropy of the eigenvalues.  The DSE is defined as the entropy of the eigenvalues of the diffusion operator $P_z$ computed on a dataset $Y$. 
$
S_D(P_z,t):=-\sum_i \alpha_{i,t} \log(\alpha_{i,t})
$
where $\alpha_{i,t}:= \frac{|\lambda_i^t|}{\sum_j |\lambda_j^t|}$,  and $\{ \lambda_i \}$ are the eigenvalues of the diffusion matrix $P_z$ and $t$ is the diffusion time. Defined this way, the DSE captures the spread of pointcloud data over its dimensions. A high DSE would correspond to an even spread across diffusion dimensions.

\vspace*{-3pt}
\paragraph{Homology and Wasserstein Distance.}
We also compute up to second order persistence homology of the hidden layer representations to assess similarity across higher homological dimensions. For a comprehensive overview of persistence homology and topological data analysis, we refer readers to \citet{Zomorodian2005}. Notably, zero-th order homology captures connected components, which, due to varying scales, may show similarities to hierarchical clustering results. Higher-order homology captures loops, voids, and additional dimensions of "holes," providing unique insights into the data's structure. 

To compare two persistence diagrams, we use the Wasserstein distance \citep{compOT}, commonly employed in topological data analysis \citep{cabanes2021wsteinTDA} due to its robustness to noise and preservation of topological features. Prominent topological features, which appear as longer intervals on persistence diagrams, are thus given more weight in this distance calculation. The Wasserstein $p$-distance between two measures \( \mu \) and \( \nu \) with finite $p$-moments is defined as:
\begin{align}
    \mathcal{W}_p(\mu, \nu) = \left( \inf_{\gamma \in \Gamma(\mu, \nu)} \int_{M \times M} d(x, y)^p \, d\gamma(x, y) \right)^{1/p}, 
\end{align}
where $ \Gamma(\mu, \nu)$ is the set of all couplings of $\mu$ and $\nu$.

\vspace*{-8pt}
\section{Results}
\label{sec:results}

\subsection{Evaluating the Manifold of Neural Networks}

 Here we compare our manifold construction technique, described in \Cref{fig:schematic}, with another more rudimentary method: computing the pairwise Frobenius norm between the raw pairwise distance matrices $D_z$ for the internal representations. This process is identical to our chosen method, except that we do not apply a Gaussian kernel onto and do not normalize the pairwise distance matrix to produce the diffusion matrix. In the second alternative method, we first construct a $k$-Nearest Neighbor graph and then compute the pairwise Frobenius norms between the resulting adjacency matrices. To assess our manifold construction, we compute the average pairwise distance between PHATE embeddings of the top $N$ highest-accuracy of $200$ MLPs trained for MNIST classification. To normalize the distances for the purpose of comparison across methods, we divide by the average pairwise distance between networks for that manifold. We observed that for each of the Top- $10$, $20$, and $30$ highest-accuracy networks, our method mapped the networks much closer together.

\begin{table}[t!]
\setlength{\tabcolsep}{20pt}
\vspace*{5pt}
    \centering
    \small
    \caption{Normalized average pairwise distances between PHATE embeddings of the $N$ highest-performing MLPs trained to classify MNIST. Distances are defined by Frobenius norms, first between diffusion matrices $P_z$, then between distance matrices $D_z$, then between $k$-NN adjacency matrices.}
    \begin{tabular}{c|c|c|c}
        \toprule
        & \textbf{Diffusion Matrices} & \textbf{Distance Matrices} & \textbf{$k$-NN Matrices}\\ 
        \hline
        \textbf{Top 10}  & 0.0790 & 0.2349 &  0.1788 \\
        \hline
        \textbf{Top 20} & 0.1347 & 0.4449 & 0.2257 \\
        \hline
        \textbf{Top 30} & 0.1373 & 0.4751 & 0.3312 \\
        \bottomrule
    \end{tabular}
    \label{tab:mlp_resnet_cnn_comparison}
\end{table}

\subsubsection{Architecture Search}

We hypothesized that hyperparameters extrapolated from models in the high-accuracy region of learned manifold could lead to high-performing models. To test this hypothesis, we selected the 30 models with the highest accuracies from our set of $200$ MLPs trained on MNIST. For these models, we identified the two most frequent combinations of weight decay and momentum, corresponding to the most common learning rate. We then trained an MLP with the same architecture using the average of these two most frequent values for weight decay and momentum and achieved a test accuracy of 97.04\% on a new set of hyperparameters. This demonstrates that our approach is a promising method for hyperparameter search. Further discussion, including how this approach may generalize across architectures, is provided in the Appendix.

\subsection{Characterizing the Landscape of Neural Networks}

We performed the characterization tests on three groups of 200 neural networks: ResNet-9s pre-trained on CIFAR-10 \citep{Krizhevsky2009LearningML} and finetuned for two epochs, MLPs trained to classify MNIST \citep{deng2012mnist}, and simple CNNs trained to classify CIFAR-10. All models had fixed initialization but varied in learning rate, momentum, and weight decay, with unique hyperparameter combinations for each model.

\paragraph{Class Structure.}

In the internal representation space, we assign each of the 100 test points a class label. We find that high accuracy correlates with larger average pairwise distances between class centroids (ResNet: $r^2=0.89$, CNN: $r^2=0.74$, MLP: $r^2=0.89$).

\begin{figure}[t!]
    \vspace*{-5pt}
    \centering
    \includegraphics[width=0.95\linewidth]{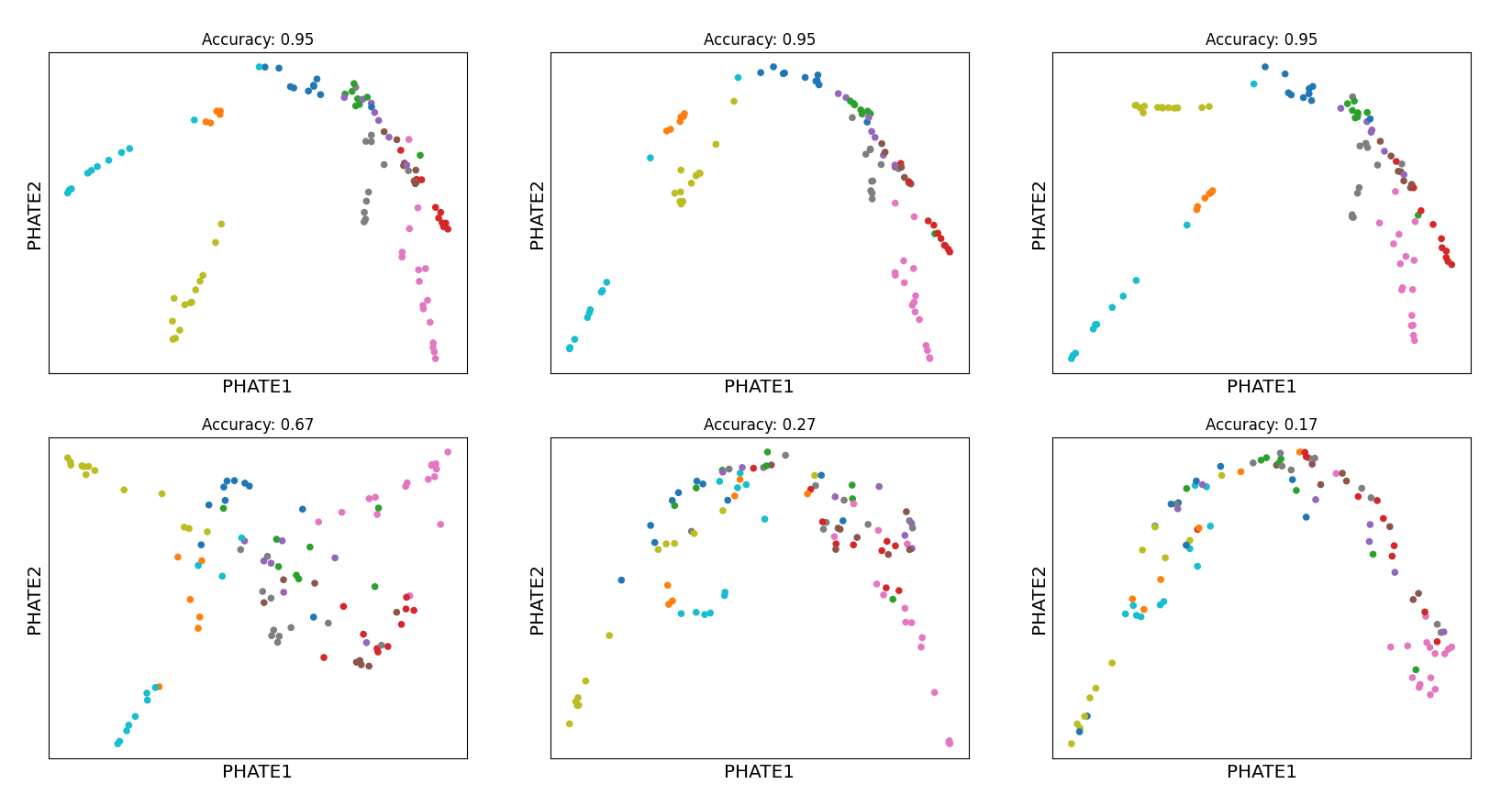}
    \caption{PHATE visualizations of embedding space for three high performing ResNets (top) and three random ResNets (bottom).}
    \label{fig:phate3good3random}
\end{figure}

\begin{figure}
    \centering
    \vspace*{-10pt}
    \includegraphics[width=0.78\linewidth]{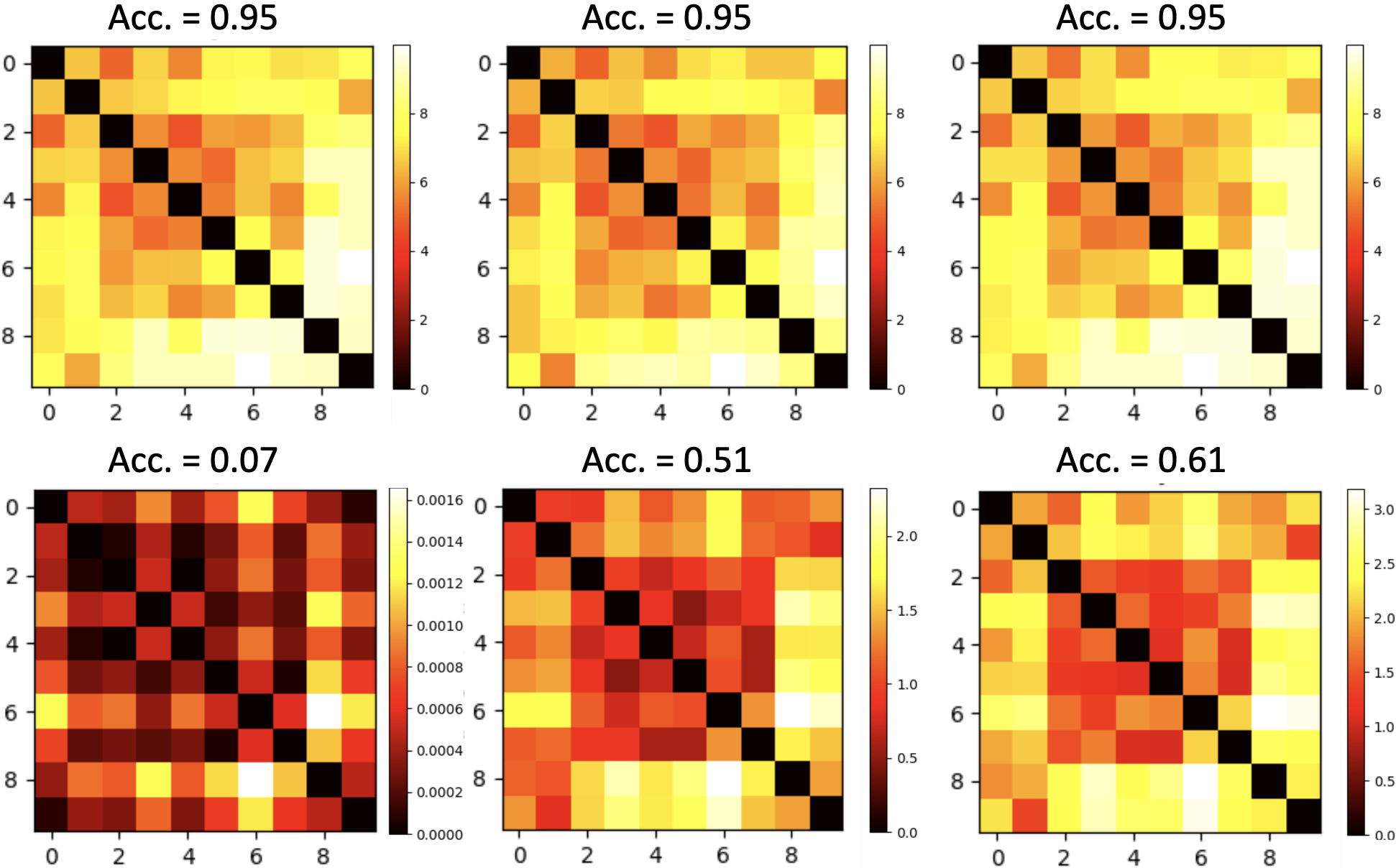}
    \caption{Class structure heatmaps for three high performing ResNets (top row) and three ResNets (bottom row). Structure for high performing models is nearly indistinguishable across all pairwise class distances. For random models, patterns are ambiguous and there is less overall separation between classes.}
    \label{fig:heatmap}
    \vspace*{-5pt}
\end{figure}

Pairwise distances between the 10 class centroids are nearly identical for higher-performing models $(A>93\%)$. This result holds despite the fact that the networks were trained on notably differing hyperparameters (learning rate \texttt{lr}$\in (0.005, 0.15)$, momentum \texttt{m}$\in(0.5,0.9)$, weight decay \texttt{wd}$\in (10^{-5}-10^{-2})$). \Cref{fig:phate3good3random} and \Cref{fig:heatmap} demonstrate that accurate networks learn similar embeddings, where the same classes are consistently structured in similar ways within the embedding space.

\begin{figure}[t!]
\centering
\includegraphics[width=0.9\linewidth]{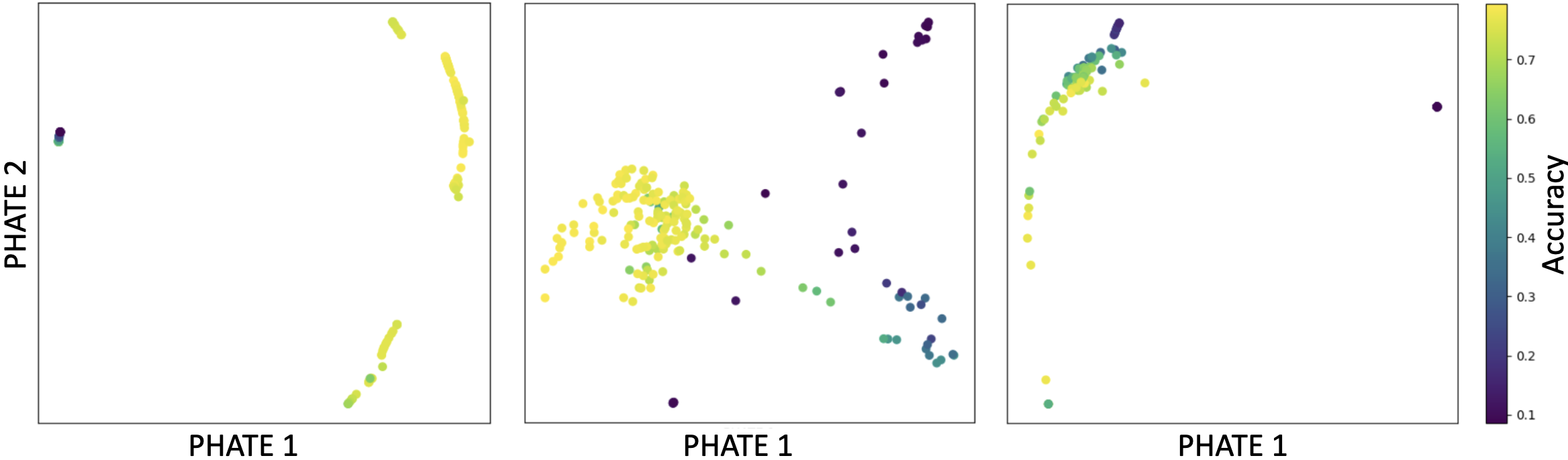}
\caption{PHATE landscapes of neural networks revealing structure by accuracy (color). \textbf{Left:} MLPs for MNIST classification. \textbf{Middle:} ResNets for CIFAR-10 classification. \textbf{Right:} CNNs for CIFAR-10 classification.}
\label{fig: default metric}
\end{figure}

We also analyze variance within classes to measure how tightly a network clusters data from the same class. Intra-class variance is comparable across the highest-performing models within each architecture. The following statistics pertain to ResNets, which exhibited the most significant accuracy range; however, similar trends were observed in CNNs and MLPs.

ResNet models were grouped into $3\%$ accuracy ranges, and for each range, we computed the average class variance and standard deviation across all classes in the dataset. The highest-performing models ($A=93\%-96\%$) demonstrated the lowest standard deviation in variance across classes (range: $0.36-0.52$, $\mu=0.42$). Models with slightly lower accuracy ($90\%-93\%$) exhibited higher variance (range: $0.76-1.03$, $\mu=0.857$), with this trend persisting as accuracy decreased. Additionally, we find that average class variance is correlated with higher neural network accuracy (ResNet: $r^2=0.84$, CNN: $r^2=0.64$, MLP:$r^2 = 0.91$). This trend can also be observed in each class in a dataset individually.

In the internal representation space, most accurate models have classes that are farther apart and less compact.     This observation is invariant to our hyperparameter adjustments and are consistent with the highest preforming models, less so as performance decreases.

\paragraph{Hierarchical Cluster Structure.}
Hierarchical clustering analysis shows whether there are structural similarities between embeddings at finer levels of granularity than under class separation. We quantify hierarchical cluster structure with the Adjusted Rand Index (ARI) and find that when computed from a single model's embedding space with respect to the ground truth, higher ARI is correlated with higher network accuracy (ResNet: $r^2=0.92$, CNN: $r^2=0.71$, MLP: $r^2 = 0.89$).

\begin{figure*}[b!]
\vspace*{-5pt}
    \centering
    \includegraphics[width=0.95\linewidth]{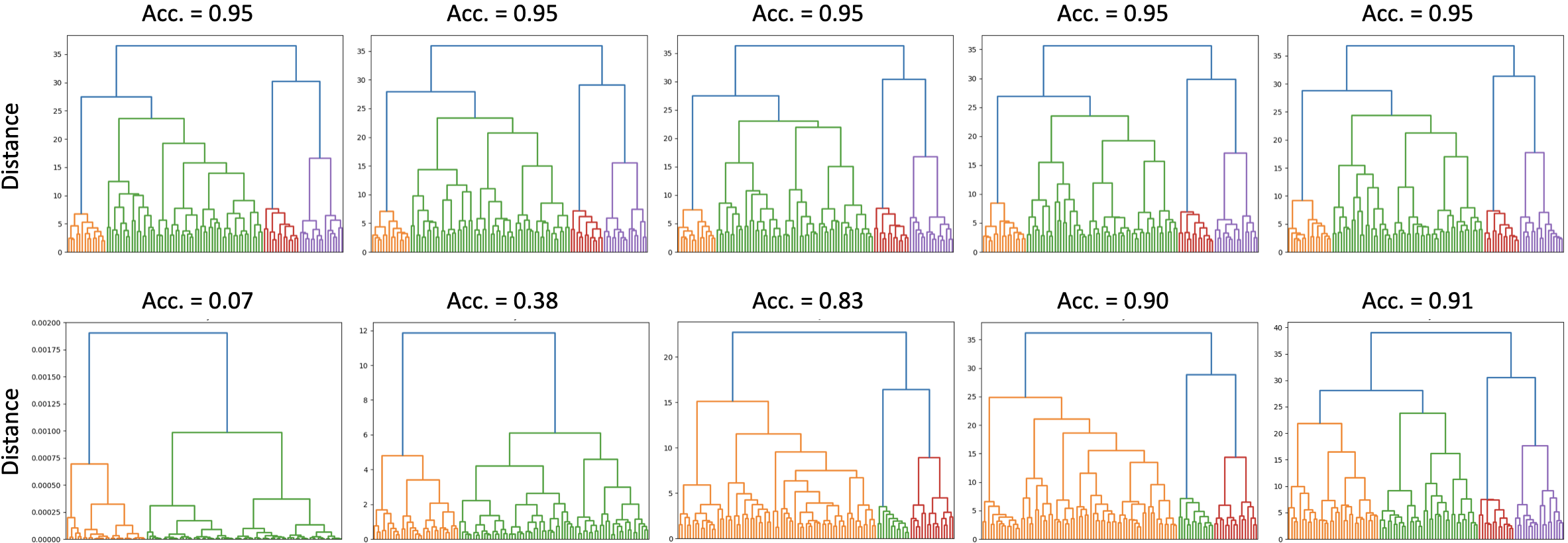}
    \caption{Dendrograms depicting microstrcuture of five high performing ResNet models (top row) and five random ResNet models (bottom row)}
    \label{fig:dendrogram}
\end{figure*}

The overall cluster structure of each model's internal representation of the same 100 test points can be visualized through dendrograms and compared numerically through pairwise ARI. For ResNets, the similarity amongst the hierarchical clustering structure of highest-performing networks is evidenced in \Cref{fig:dendrogram}.  We observe very similar overall  ``microstructure" between highest-performing models.  These patterns are invariant across hyperparameters and the similarity in microstructure certainly does not hold as accuracy decreases as evidenced in the bottom row of the figure. Pairwise ARI between models is higher for better performing models than that of moderate and poorly-performing models. Using the $3\%$ accuracy batches from earlier we find a moderate correlation between accuracy range midpoint and average pairwise ARI (ResNet: $r^2=0.60$, CNN: $r^2=0.52$) and a strong correlation for the simplest class of networks (MLP: $r^2 = 0.97$).
Our analysis shows that the highest-performing models cluster data closer to ground truth labels and have similar microstructures, while moderate and poorly performing models are more divergent.

\begin{figure}[t!]
    \vspace*{-5pt}
    \noindent
    \centering
    \includegraphics[width=0.9\linewidth]{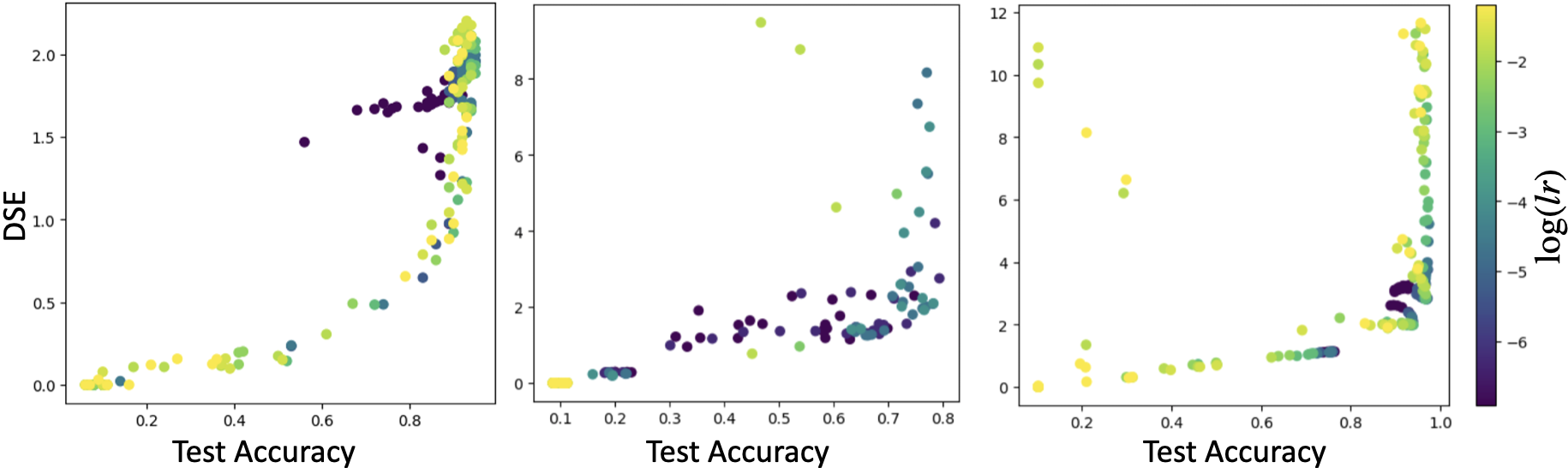}
    \caption{Diffusion Spectral Entropy increases with test accuracy across the three architectures (MLP, ResNet, CNN). Points are colored by learning rate.}
    \label{fig:DSE}
    \vspace*{-8pt}
\end{figure}

\paragraph{Diffusion Spectral Entropy} High-performance networks across architectures tend to have higher Diffusion Spectral Entropy (DSE) associated with their internal representations, using $t=0$ because of the low noise in the dataset. \Cref{fig:DSE} indicates this positive relationship between network performance and spread across diffusion matrix dimensions.

These results indicate that high-performing networks exhibit more pronounced cluster structures in their hidden layer embeddings. This pattern is in agreement with the findings of \citet{liao2023assessingneuralnetworkrepresentations}: for $k$ separated clusters, they prove that as $t$ grows to infinity, $S_D(P_z, t)$ tends to $\text{log}(k)$. This makes sense because for graphs that have disconnected components, we would expect to have multiple eigenvectors with non-trivial eigenvalues, or a more even spread of these eigenvalues across the dimensions represented by the eigenvectors. Therefore, we can intepret DSE as measuring the number of significant eigendirections in the data. On the other hand, if the dataset only forms one uniform cluster, since it can be represented by a fully connected graph (i.e. an ergodic process), we would expect the DSE to approach $0$ as $t$ tends to infinity since we would have only one non-trivial eigendirection. In other words, the diffusion matrix that represents such data would approach a rank 1 steady-state matrix.

\begin{wrapfigure}{r}{0.34\textwidth}
\vspace*{-12pt}
    \centering
    \includegraphics[width=\linewidth]{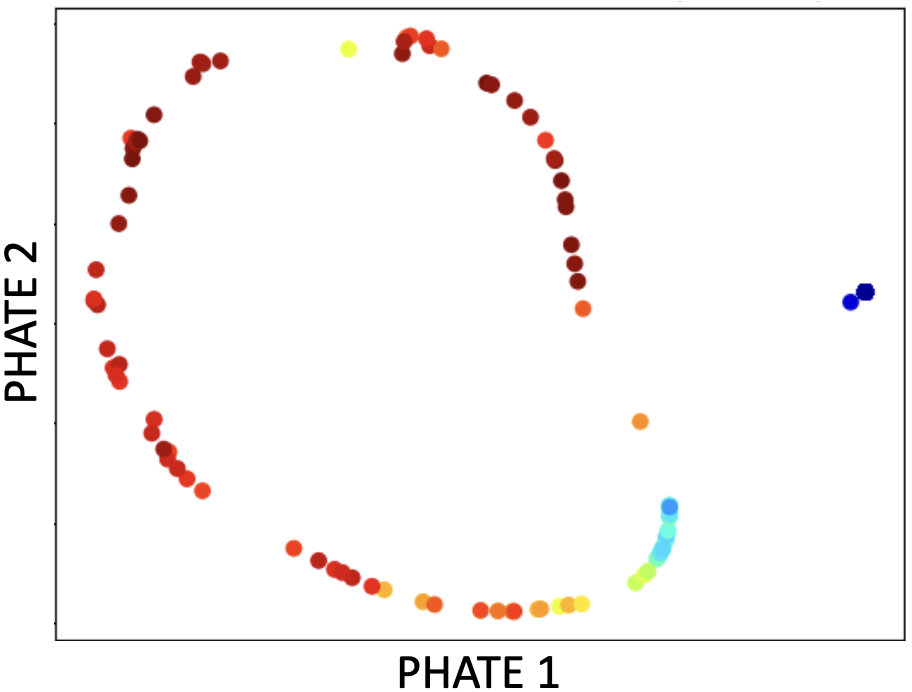}
    \caption{PHATE plot of 200 CNNs colored by their accuracies. Wasserstein Distance between persistence diagrams are used as the distance metric.}
    \label{fig:PDCNN2}
    \vspace*{-25pt}
\end{wrapfigure}

\paragraph{Persistent Homology} Across architectures, we have found that topological features can characterize neural network accuracy, independent of hyperparameters, in the sense that Wasserstein distances between persistence diagrams of internal representations are smaller for high preforming networks. Using PHATE to visualize these Wasserstein distances, \Cref{fig:PDCNN2} illustrates our observed result for 200 Convolution Neural Networks, coloring by accuracy.

\begin{figure}[t!]
    \noindent
    \centering
    \includegraphics[width=0.95\linewidth]{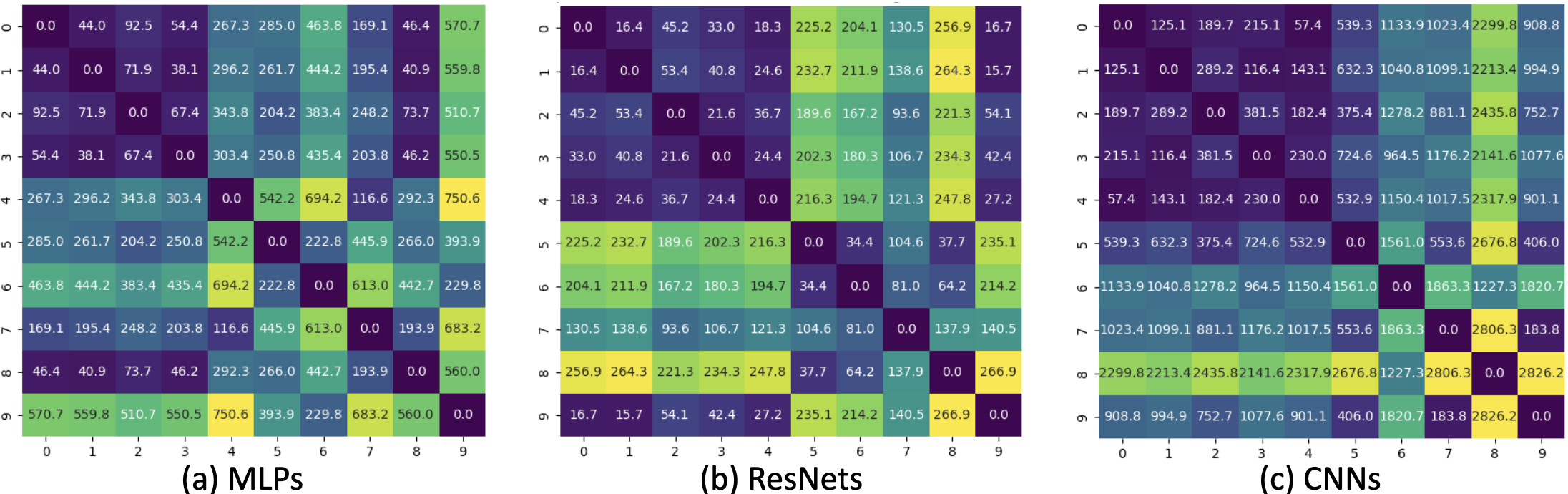}
    \caption{Heat maps for Wasserstein distance between persistence diagrams of five high performing networks (indices 0-4) and five random networks (indices 5-9).}
    \label{fig:TDAHM}
\end{figure}

\Cref{fig:TDAHM} summarizes these findings, illustrating through distance heatmaps how accuracy shapes topology across each architecture observed. This result aligns closely with the structural characterization findings above, suggesting that the way a successful neural network clusters its data is not only geometrically but also topologically identifiable and remains consistent across varying hyperparameters.

Overall, our results indicate that high-performing networks exhibit similar representations with comparable micro and macro scale structures, as quantified by both persistent homology and hierarchical clustering. Additionally, these networks display greater class separation, increased within-class variance, and generally higher entropy. This suggests that such embeddings spread data points in meaningful patterns across diffusion dimensions, capturing informative structural relationships.

\section{Conclusion}

In conclusion, we introduced a novel framework for exploring and understanding the landscape of neural networks by examining their internal representations, with each representation corresponding to a single network. This approach allowed us to define a distance metric between these internal representations, which we then used to embed networks into a low-dimensional manifold. By characterizing the manifold with features such as hierarchical clustering, spectral entropy, and topological structure, we revealed that high-performing networks consistently exhibit similar internal representations.

\section*{Acknowledgments}

This project was conducted as part of the Summer Undergraduate Math Research at Yale (SUMRY) REU program, supported by NSF DMS (Award 2050398). D.B. acknowledges support from the Yale-Boehringer Ingelheim Biomedical Data Science Fellowship and the Kavli Institute for Neuroscience Postdoctoral Fellowship. I.A. and S.K. acknowledge support from the NSF DMS Grant (Award 2327211). Additionally, S.K. is supported by the NSF CAREER Award (IIS 2047856) and the NSF MEDIUM Grant (IIS 2403317).

\clearpage

\printbibliography

\clearpage

\clearpage

\appendix

\crefalias{section}{appsec}
\crefalias{subsection}{appsec}
\crefalias{subsubsection}{appsec}

\setcounter{equation}{0}
\renewcommand{\theequation}{\thesection.\arabic{equation}}

\onecolumn

\section*{\Huge Appendix}
\label{sec:appendix}

\vspace*{10pt}

\section{Datasets for the Manifold of Neural Networks}
To compare methods for defining the manifold of neural networks, the manifold was computed and visualized for three classes of 200 neural networks: ResNet-9s pre-trained to classify images from the CIFAR-10 dataset and finetuned for two additional epochs, Simple MLPs trained for MNIST classificatioj and Simple CNNs trained for CIFAR-10 classification. Models all had fixed weight initialization but spanned a range of learning rate, momentum, and weight decay with each model having a unique combination of hyperparameters. Below we outline our method using the MNIST dataset as an example, but we used the same set of hyperparameters for stochastic gradient descent optimization on ResNets and CNNs. Code for each is provided in a GitHub repository. 

The architecture of the network trained to classify MNIST is an input layer with dimension 784, followed by two fully connected hidden layers with dimension 50 and 20, respectively. The activation function used for both hidden layers is ReLU. The output layer contains 10 neurons, with the softmax activation function. The 200 models for each architecture and task exhausted all combinations of the following hyperparameters: $\text{Learning Rates} \in \{0.001, 0.005, 0.01, 0.05, 0.1, 0.15, 0.2, 0.3\}$ $\text{Momentum} \in \{0.5, 0.6, 0.7, 0.8, 0.9\}$ $\text{Weight Decay} \in \{1 \times 10^{-5}, 1 \times 10^{-4}, 1 \times 10^{-3}, 1 \times 10^{-2}, 0.1\}$. 

\paragraph{Machines}
Our networks were trained using Google Colab Pro (ResNets), NVIDIA a100 GPU (MLPs), and GeForce GTX 1080Ti GPUs (CNNs).

\section{Manifold Method Comparisons}
\paragraph{Alternative 1 - Euclidean Distance}
Let $z \in [Z]$ be a neural network. Similarly as in section 4.1, we concatenate internal representations $a_z^k$  $\forall k \in [Y]$ to obtain $E_z$ and its $n \times n$ pairwise distance matrix $D_z$. Here, $D_z$ is our characteristic embedding of neural network $z$.

We repeat this embedding process $\forall z \in [Z]$. Using these embeddings, we compute the pairwise Frobenius norm between $\{(E_i,E_j) : i,j \in [Z]\}$ for each trained neural network $z$ to obtain the $m \times m$ distance matrix $N_1$, where $m = |Z|$. PHATE was then used to visualize $N_1$. Note that this process is identical to our chosen metric, except for neglecting to pass the pairwise distance matrix into a Gaussian kernel. 

\begin{figure}[t!]
    \centering
    \includegraphics[width=0.9\linewidth]{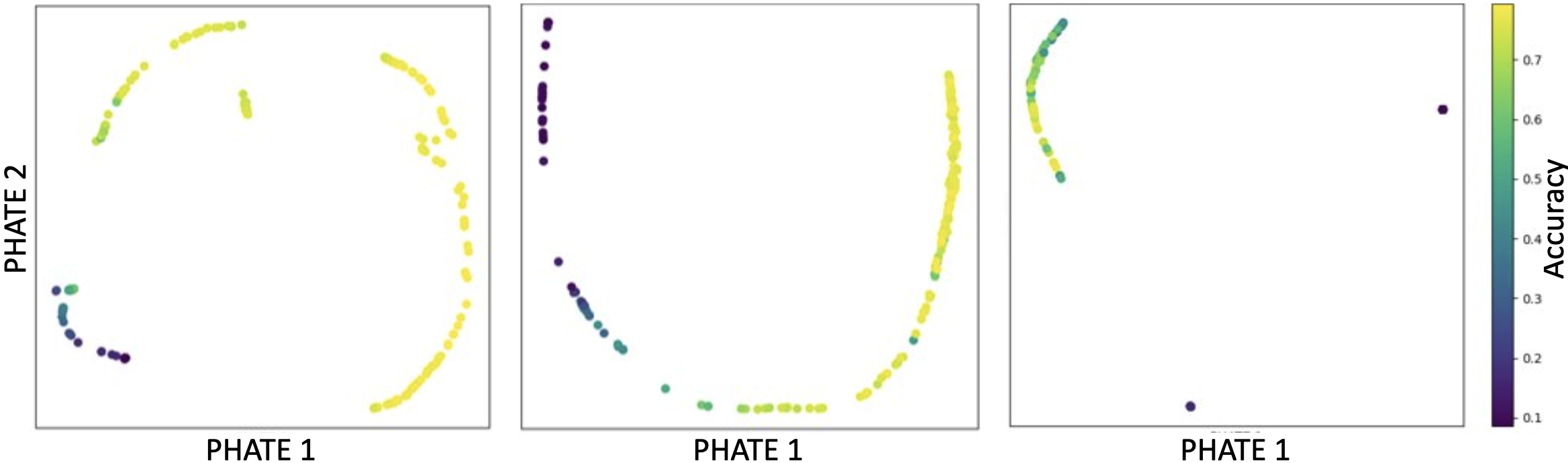}
    \caption{Landcape of neural networks by PHATE using only Euclidean distance. \textbf{Left:} MLPs for MNIST classification. \textbf{Middle:} ResNets for CIFAR-10 classification. \textbf{Right:} CNNs for CIFAR-10 classification.}
    \label{fig: euclidean metric}
\end{figure}

In this alternative, the distance between neural networks is defined by the similarity or dissimilarity of their pairwise Euclidean matrix $E_z$. The resulting landscape can be seen in \Cref{fig: euclidean metric}.

\paragraph{Alternative 2 - Comparing $k$-NN Graphs} We repeat the same steps as above to obtain $E_z$ and $D_z$. Then, use the $k$-Nearest Neighbor algorithm, with $k$ = 5, to construct a $k$-NN similarity matrix $W_z$. Here, $W_z$ is our characteristic embedding of neural network $z$.

We repeat this embedding process $\forall z \in [Z]$. Using these embeddings, we compute the pairwise Frobenius norm between $\{(E_i,E_j) : i,j \in [Z]\}$ for each trained neural network $z$ to obtain the $m \times m$ distance matrix $N_2$. PHATE was then used to visualize $N_2$. 

\begin{figure}[t!]
    \centering
    \includegraphics[width=0.9\linewidth]{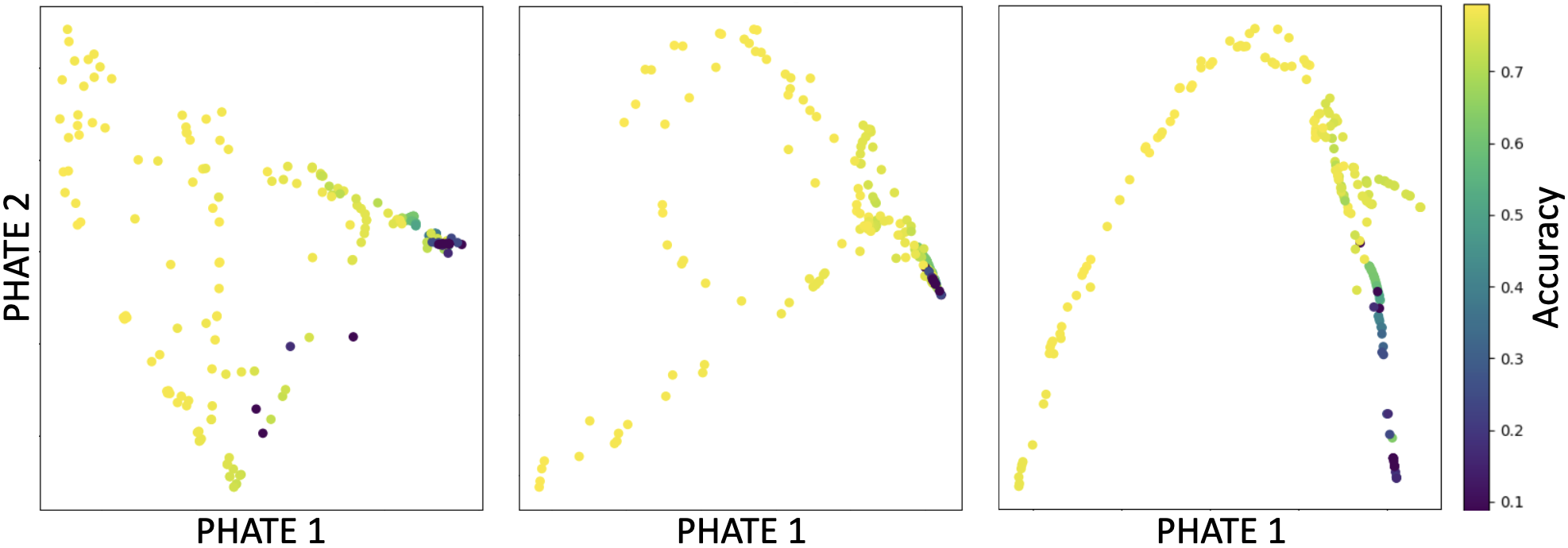}
    \caption{Landscape of neural networks by PHATE using distance between the (left to right) first, second, and third layer weight matrices of the MLPs.}
    \label{fig: weight metric}
\end{figure}

\paragraph{Alternative 3 - Comparing Weight Matrices} Let $z \in [Z]$ be a neural network. Its output after the $k^{th}$ layer is defined by $h_k = f_k(W^{(k)} + b^{(k)})$. We define our distance matrix, $D_z$, across the manifold, such that $D_{i,j} = \| W_i^{(k)} - W_j^{(k)} \|_F $, where $F$ is the Frobenius norm. The distance matrix is then visualized using PHATE. Here, the distance between neural networks is defined by the similarity or dissimilarity of their weight matrix. The resulting landscape can be seen in \Cref{fig: weight metric}. Using the evaluation metric from \Cref{tab:mlp_resnet_cnn_comparison}, the weights-based manifold places the normalized average distance between Top-N highest accuracy models at: Top 10: 0.3721, Top 20: 0.5316, and the Top 30: 0.5504. 

\section{TDA}

\begin{figure}[h!]
    \noindent
    \centering
    \includegraphics[width=0.9\linewidth]{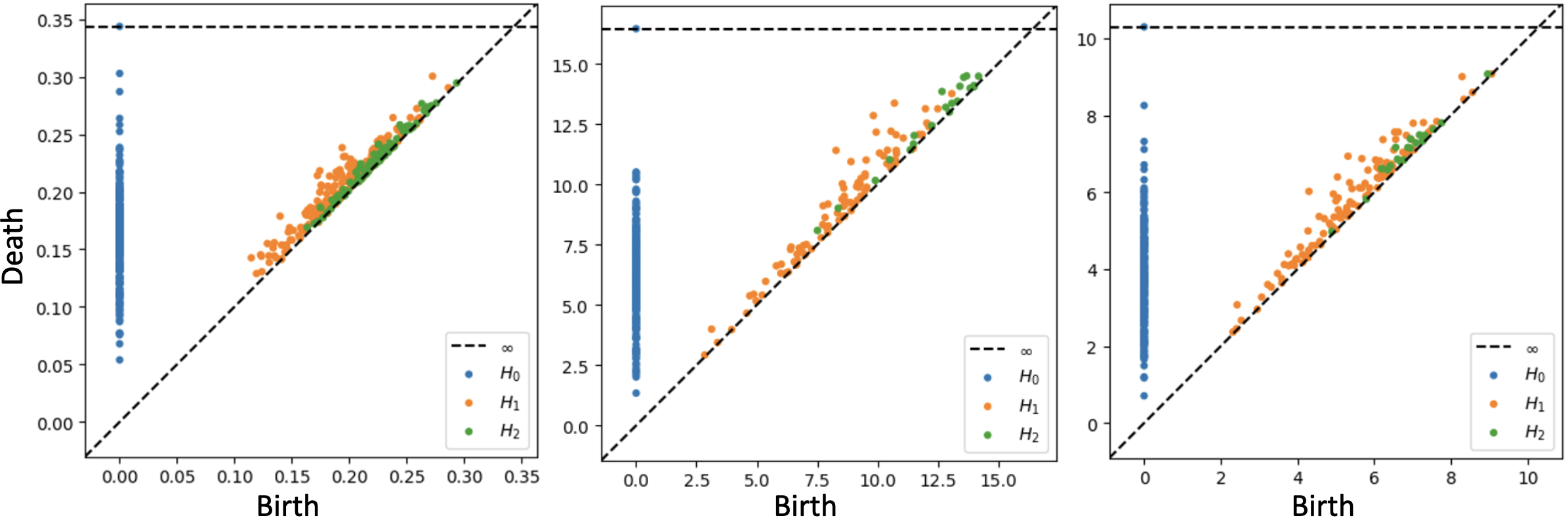}
    \caption{Persistence diagrams representing three trained MLPs with different accuracies (\textbf{Left} : 0.1, \textbf{Middle} : 0.97, \textbf{Right} : 0.97). The $H_0$ in high-performing networks survive longer than those in the low-performing network, illustrating the improved seperation of clusters.}
    \label{fig:PDMLP}
\end{figure}

\Cref{fig:PDMLP} presents one low-performing and two high-performing MLPs, all trained on their unique set of hyperparameters. Similarities in persistence diagrams are visually clear, despite each network being unique. 

\clearpage

\section{Diffusion Spectral Mutual Information} The Shannon mutual information between two variables measures how much information one captures about the other, or the decrease in the entropy from knowing one variable. The mutual information between two random variables is zero if and only if they are independent. It is defined as 
$
I(X; Y) = H(X) - H(X|Y) 
        = H(X) - \sum_i p(Y = y_i) H(X | Y = y_i)
$
Based on the previous definition of DSE, we can now define Diffusion Spectral Mutual Information (DSMI) as 
$
I_D(Z; Y) = S_D(P_z,t) - \sum_{y_i \in Y} p(Y=y_i) S_D(P_{Z|Y=y_i}, t)
$
where ${P}_{Z|Y=y_i}$ is the transition matrix computed on the subset of $Z$ that has class label $Y$.

The DSMI allows us to compare the distribution of eigenvalues of two diffusion matrices. In this paper, we limit ourselves to this definition and only compute mutual information between the penultimate layer embeddings and the output of the network, or the true labels. 

\begin{figure}[ht]
    \noindent
    \centering
    \includegraphics[width=0.9\linewidth]{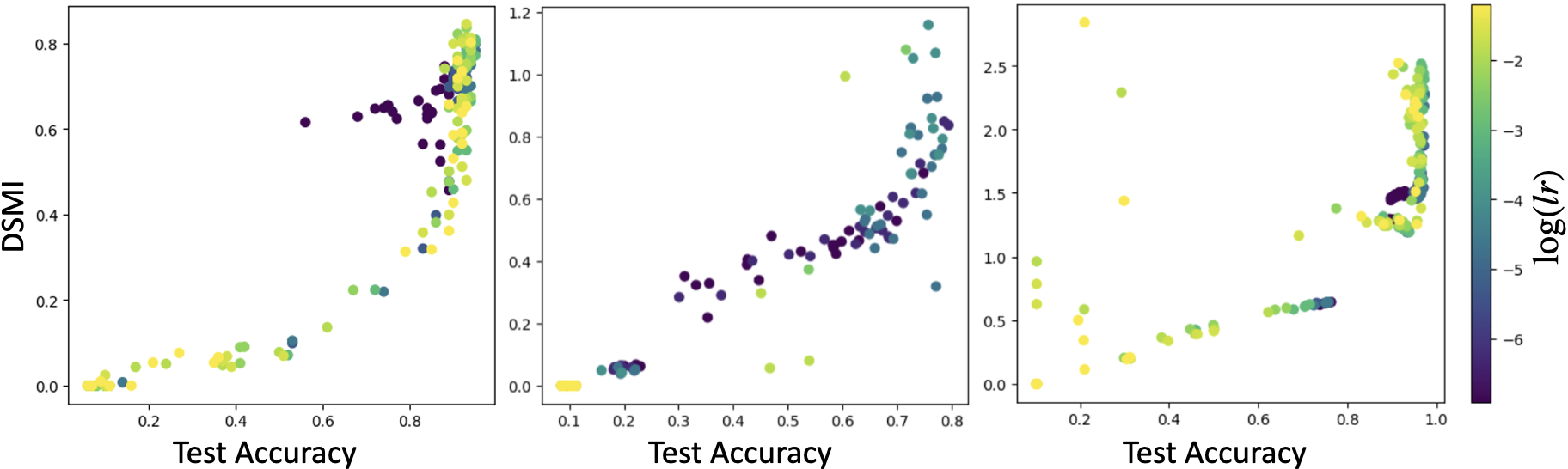}
    \caption{Plots of DSMI between penultimate layer embeddings and true class labels, versus test accuracy. The points are colored by learning rate.}
    \label{fig:DSMI}
\end{figure}

We also find that neural networks with high test accuracy demonstrate higher diffusion mutual information between their penultimate layer embedding and the true class labels (\Cref{fig:DSMI}), across all three architectures. This is not a surprising result, since we would expect good neural networks to already showcase cluster structure at the penultimate layer. In other words, one would expect the penultimate layer embeddings to be very similar to the output of the network.

\begin{figure}[t!]
    \noindent
    \centering
    \includegraphics[width=0.8\linewidth]{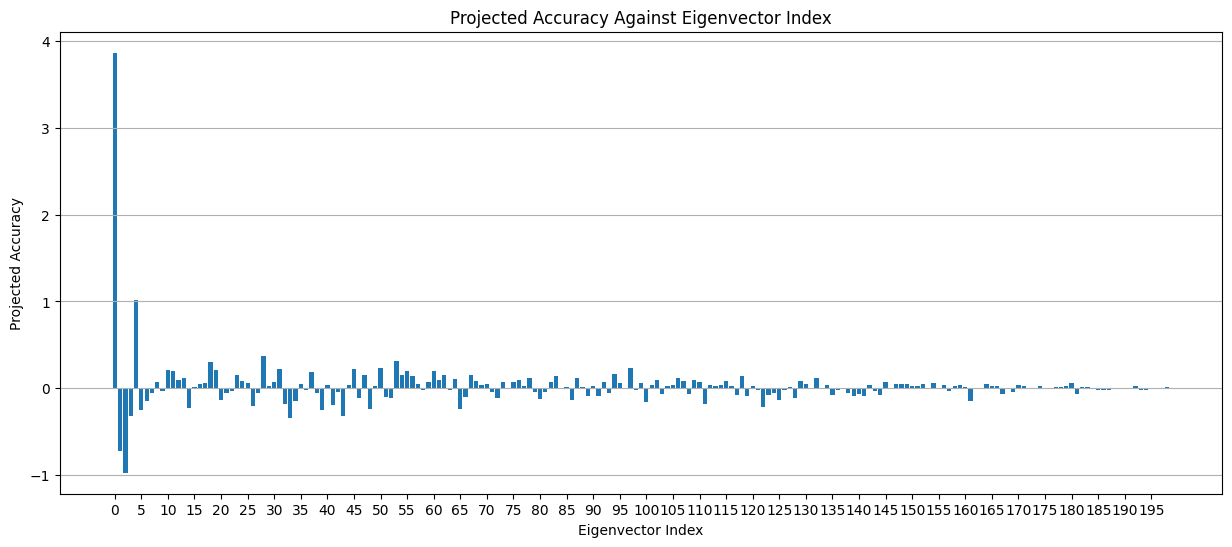}
    \includegraphics[width=0.8\linewidth]{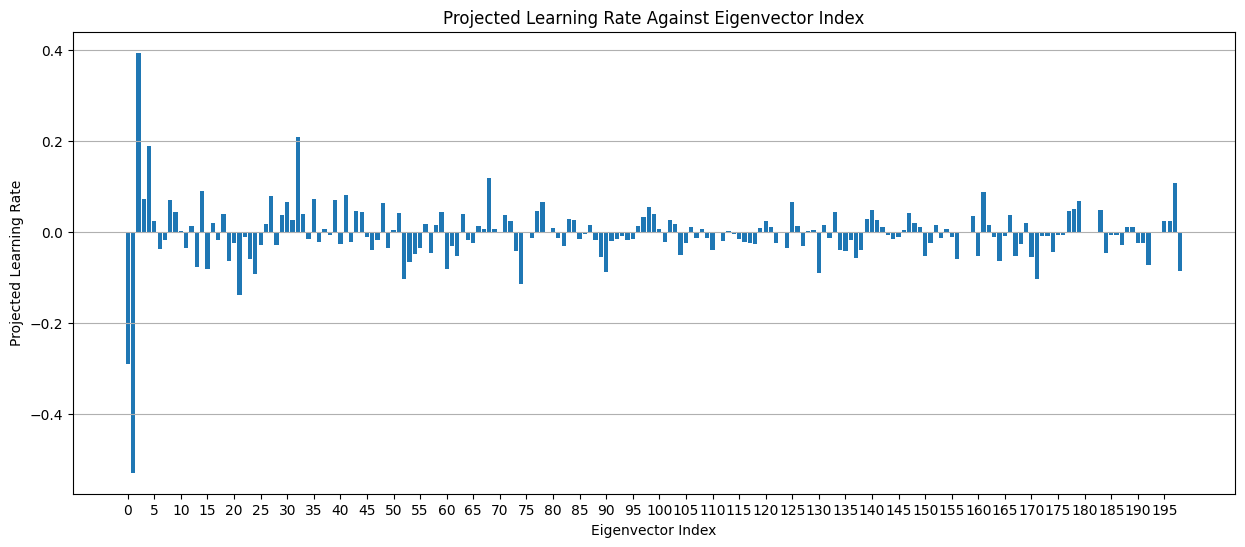}
    \includegraphics[width=0.8\linewidth]{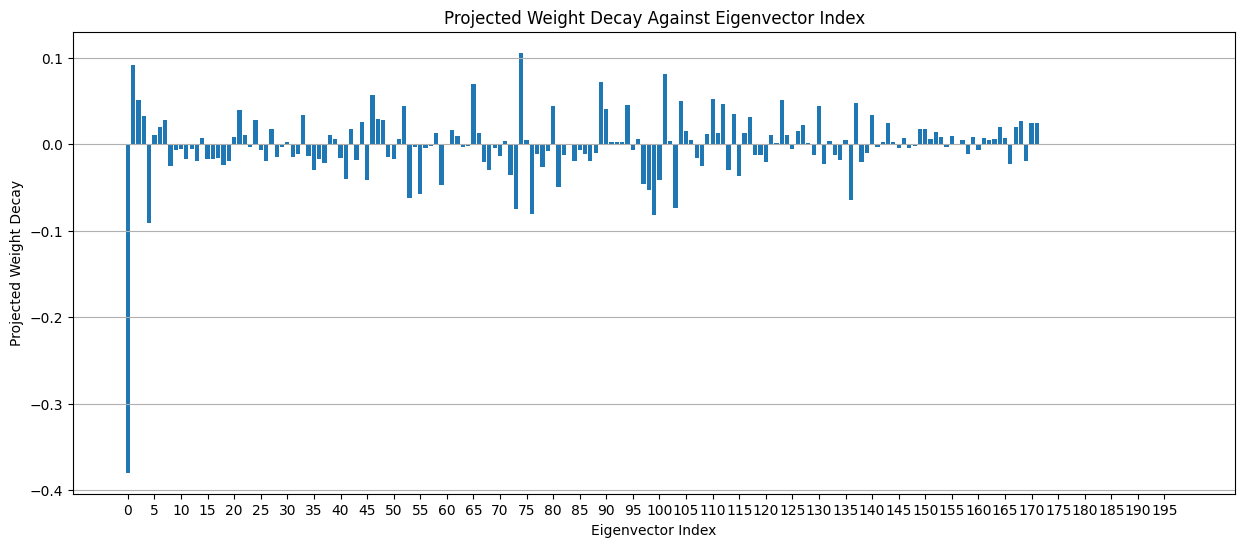}
    \includegraphics[width=0.8\linewidth]{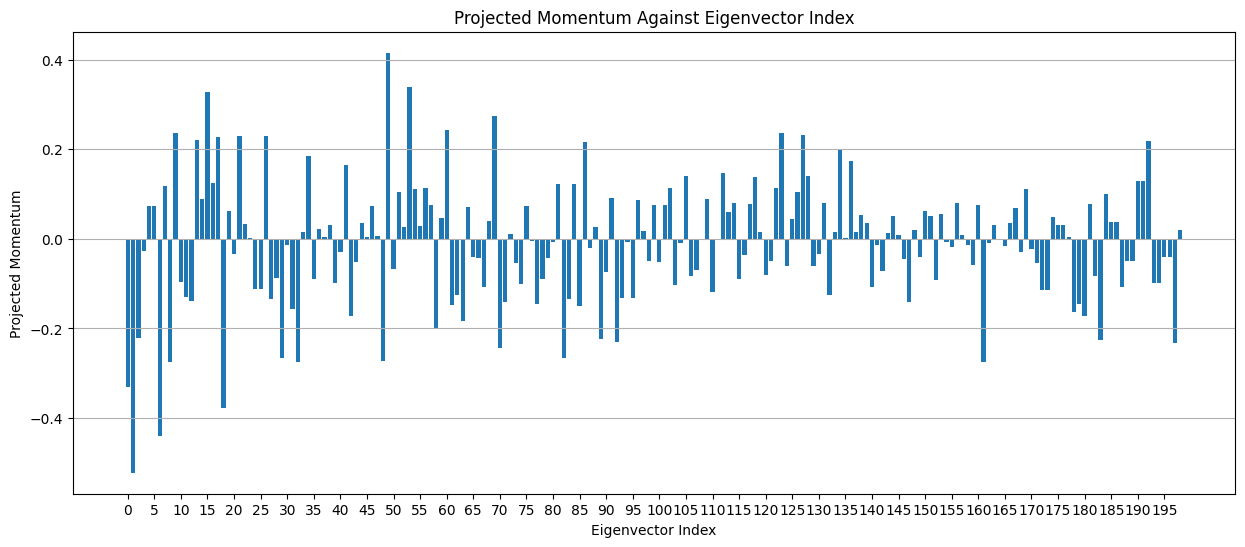}
    \caption{Plots of inner product between signals (top to bottom: \textbf{a)} accuracy, \textbf{b)} learning rate, \textbf{c)} weight decay, \textbf{d)} momentum) and the Laplacian's eigenspectrum for the manifold of ResNets. The trivial eigenvector is omitted. We compute the Graph Laplacian quadratic form to provide a measure for smoothness: \textbf{a)} $1335.4$, \textbf{b)} $800.3$, \textbf{c)} $388.1$, \textbf{d)} $1525.5$.}
    \label{fig:resnet_fourier}
\end{figure}

\begin{figure}[t!]
    \noindent
    \centering
    \includegraphics[width=0.8\linewidth]{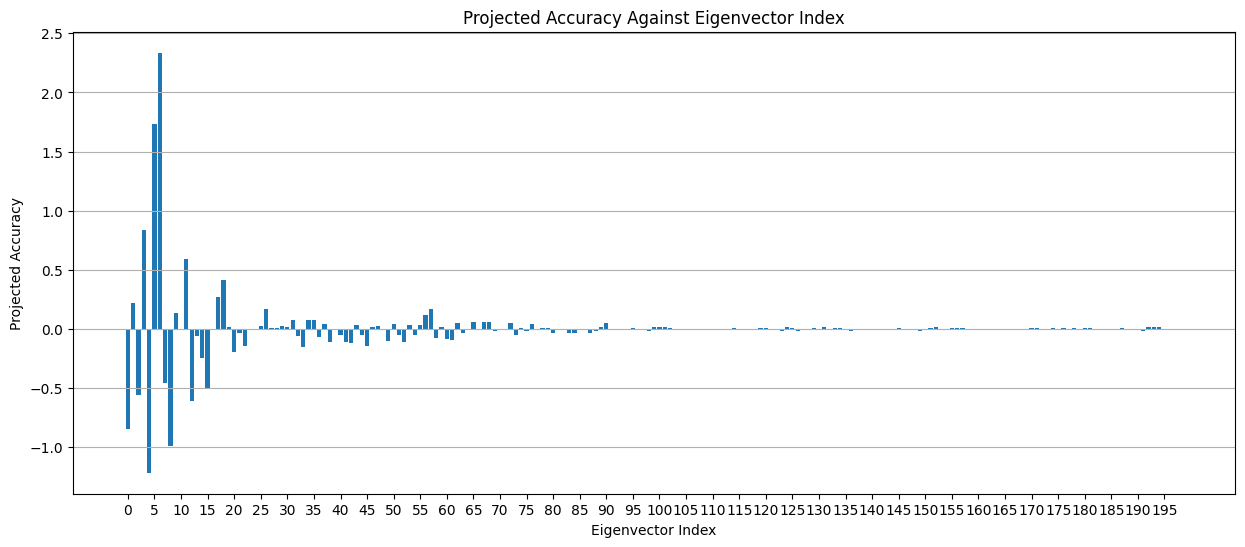}
    \includegraphics[width=0.8\linewidth]{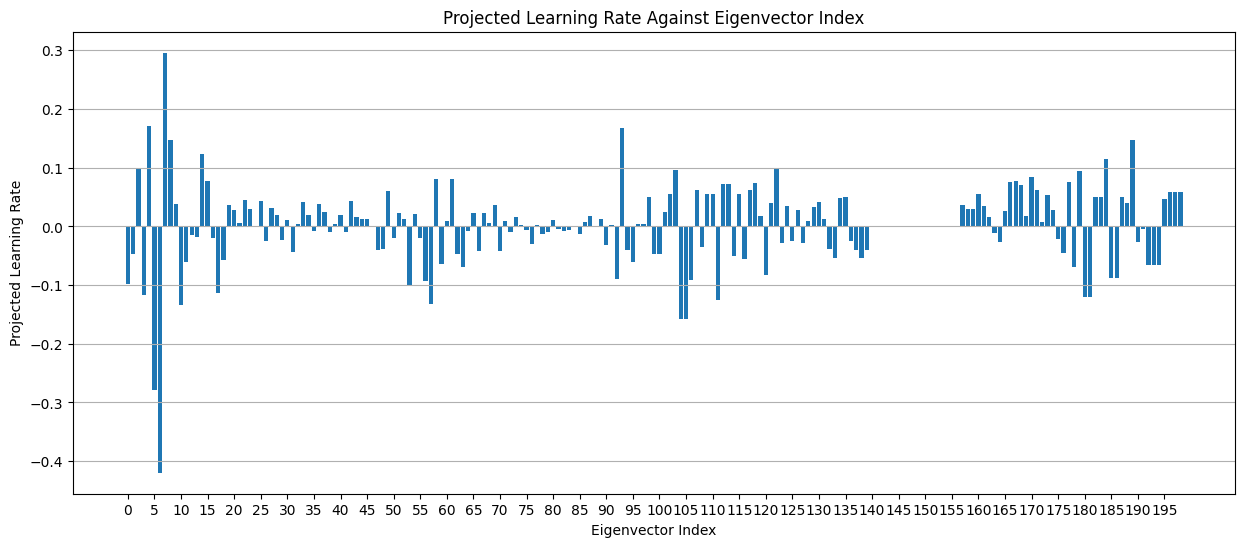}
    \includegraphics[width=0.8\linewidth]{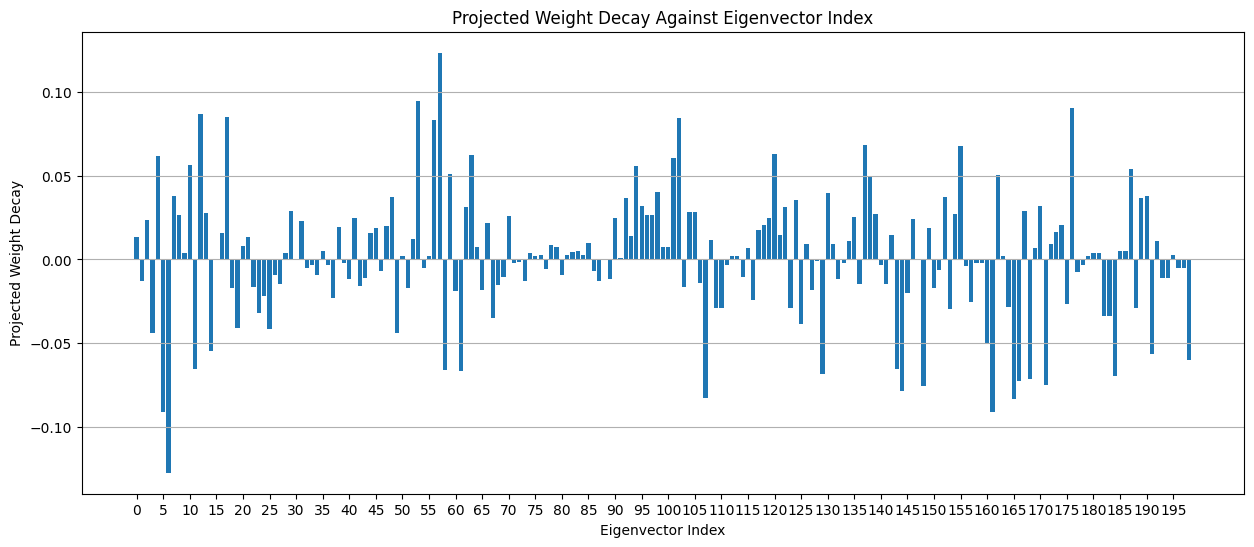}
    \includegraphics[width=0.8\linewidth]{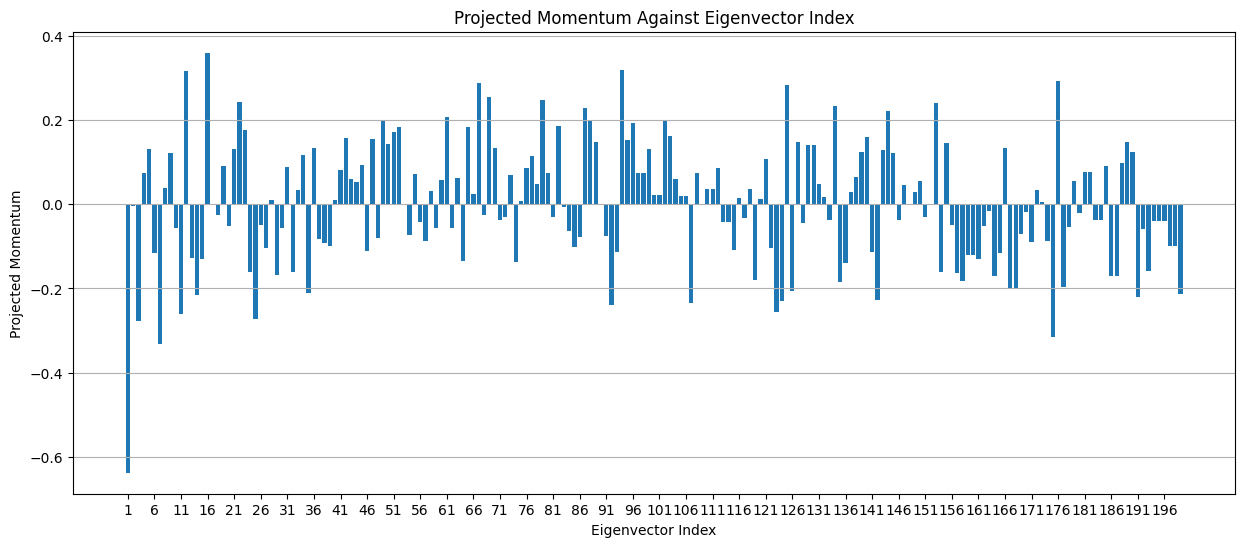}
    \caption{Plots of inner product between signals (top to bottom: accuracy, learning rate, weight decay, momentum) and the Laplacian's eigenspectrum for the manifold of CNNs. We compute the Graph Laplacian quadratic form to provide a measure for smoothness: \textbf{a)} $891.8$, \textbf{b)} $688.5$, \textbf{c)} $384.1$, \textbf{d)} $1497.5$.}
    \label{fig:cnn_fourier}
\end{figure}

\paragraph{Graph Fourier Transform} For each architecture, we generate a $n \times n$ affinity matrix $W$, based on the Frobenius norms between penultimate layer diffusion matrices. We look at the eigenspectrum of the Laplacian $L = WD^{-1}$ of the affinity matrix (where $D$ is the degree matrix). We compute Graph Fourier Transform (GFT) \citet{Shuman_2013} for this matrix. We analyze the inner product between the Laplacian eigenspectrum computed from these harmonics and scalar signals on the graph, which we define as signals
$\in$ \{\texttt{accuracy, learning rate, weight decay, momentum}\}. Essentially, we use the GFT of the neural manifold graph to understand which properties of the neural networks are encoded as low vs. high frequency on the manifold (\Cref{fig:resnet_fourier,fig:cnn_fourier}).

\begin{itemize}
    \item \textbf{Accuracy}: We observe that across different architectures, accuracy emerges as a low-frequency signal. This indicates that the manifold we're constructing effectively organizes neural networks by their performance. The implication is that our approach to building the manifold is appropriate and beneficial for identifying good hyperparameter ranges.
    \item \textbf{Learning Rate}: Learning rate is also a low-frequency signal. This implies there is a region where we can interpolate the learning rate effectively, providing justification for the selected hyperparameter values and highlighting ranges that are worth exploring.
    \item \textbf{Momentum and Weight Decay}: These are high-frequency signals. This indicates more variability and suggests these parameters might require more granular tuning and cannot be easily interpolated.
\end{itemize}

\clearpage

\clearpage

\begin{figure}[H]
    \centering
    \includegraphics[width=0.9\linewidth]{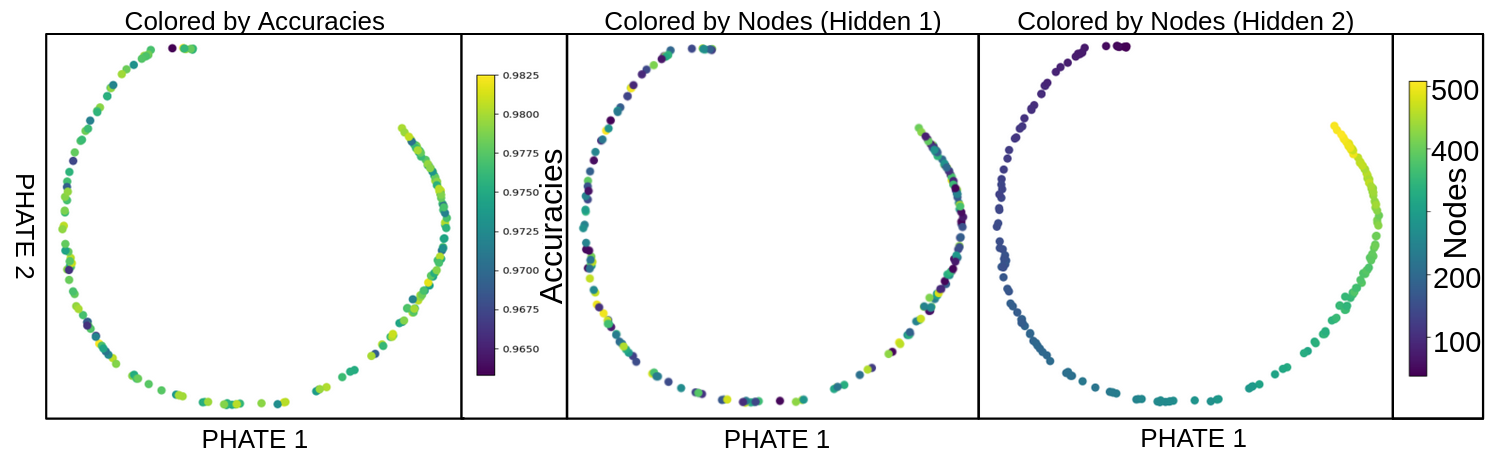}
    \caption{PHATE Plot of Penultimate Embeddings. \textbf{From left to right}: Colored by Accuracy, Colored by Number of nodes in the first hidden layer, Colored by Number of nodes in the second hidden layer.}
    \label{fig:3plotsEmbedding}
\end{figure}

\paragraph{Architecture Search} We trained 200 neural networks on MNIST. The architecture consisted of an input layer, two hidden layers, and an output layer. The number of nodes in both hidden layers was randomly chosen between 32 and 512. We extracted each model's internal representation (penultimate embeddings) of the test set and constructed a $200 \times 200$ pairwise distance matrix based on these embeddings. We visualized this matrix using PHATE (\Cref{fig:3plotsEmbedding}).

\begin{figure}[H]
    \centering
    \includegraphics[width=0.9\linewidth]{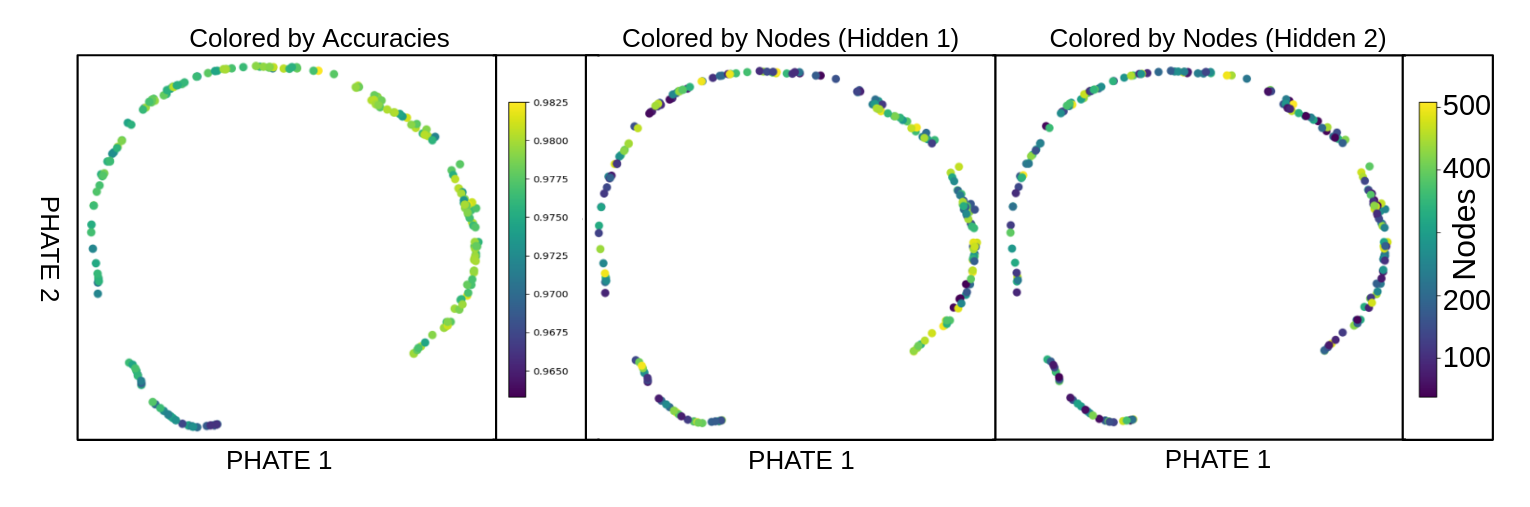}
    \caption{PHATE Plot of Distance Matrices from Penultimate Embeddings, \textbf{From left to right}: Colored by Accuracy, Colored by Number of nodes in the first hidden layer, Colored by Number of nodes in the second hidden layer.}
    \label{fig:3plotsDistance}
\end{figure}

For each internal representation, we computed an individual pairwise distance matrix, yielding 200 square matrices with dimensions equal to the size of the MNIST test set. From these individual distance matrices, we formed a single $200 \times 200$ pairwise distance matrix by computing the Frobenius norm between every pair of distance matrices. This new pairwise distance matrix was then visualized using PHATE (\Cref{fig:3plotsDistance}).

These figures indicate that the number of nodes of the second hidden layer is a low-frequency signal on the manifold of neural networks. These findings indicate that there is a region in the manifold on which we can interpolate the second-layer size effectively, shoring the ranges that are worth exploring.  

\end{document}